%% file: main_arxiv.tex
\documentclass{article}

\usepackage{ijcai24}

\usepackage{times, kotex}
\usepackage{soul}
\usepackage{url}
\usepackage[hidelinks]{hyperref}
\usepackage[utf8]{inputenc}
\usepackage[small]{caption}
\usepackage{graphicx}
\usepackage{amsmath}
\usepackage{amsthm}
\usepackage{booktabs}
\usepackage{algorithm}
\usepackage{algorithmic}
\usepackage[switch]{lineno}
\newrobustcmd{\B}{\bfseries}
\usepackage{mathtools}
\usepackage{amsfonts}
\usepackage{xcolor}
\usepackage{color,colortbl}

\title{Disrupting Diffusion-based Inpainters with Semantic Digression}

\author{
Geonho Son$^{*}$$^1$,
Juhun Lee$^{*}$$^1$,
\And
Simon S. Woo$^{1,2}$
\affiliations
$^1$Department of Artificial Intelligence
\\
$^2$Computer Science \& Engineering Department \\
Sungkyunkwan University\\
\emails
\{sohn1029, josejhlee, swoo\}@g.skku.edu
}

\date{January 2024}

\begin{document}

\maketitle

\input{contents/0_abstract}
\input{contents/1_introduction}

\input{contents/2_related_work}

\input{contents/2_1Background}
\input{contents/3_method}
\input{contents/4_experiment}
\input{contents/5_Discussion}
\input{contents/6_conclusion}

\bibliographystyle{named}
\bibliography{biblio}

\clearpage
\appendix

\input{contents/appendix}

\end{document}

%% file: contents/0_abstract.tex
\begin{abstract}

The fabrication of visual misinformation on the web and social media has increased exponentially with the advent of foundational text-to-image diffusion models. Namely, Stable Diffusion inpainters allow the synthesis of maliciously inpainted images of personal and private figures, and copyrighted contents, also known as deepfakes. To combat such generations, a disruption framework, namely Photoguard, has been proposed, where it adds adversarial noise to the context image to disrupt their inpainting synthesis. While their framework suggested a diffusion-friendly approach, the disruption is not sufficiently strong and it requires a significant amount of GPU and time to immunize the context image. In our work, we re-examine both the minimal and favorable conditions for a successful inpainting disruption, proposing DDD, a “$\textbf{\small{D}}$igression guided $\textbf{\small{D}}$iffusion $\textbf{\small{D}}$isruption” framework. First, we identify the most adversarially vulnerable diffusion timestep range with respect to the hidden space. Within this scope of noised manifold, we pose the problem as a semantic digression optimization. We maximize the distance between the inpainting instance's hidden states and a semantic-aware hidden state centroid, calibrated both by Monte Carlo sampling of hidden states and a discretely projected optimization in the token space. Effectively, our approach achieves stronger disruption and a higher success rate than Photoguard while lowering the GPU memory requirement, and speeding the optimization up to three times faster.
\end{abstract}

\begin{figure}[ht]
\centering
\includegraphics[width=\columnwidth]{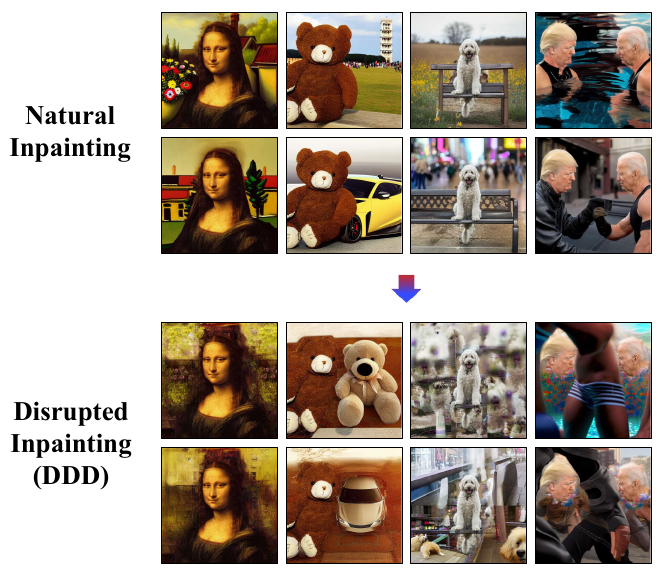}
\caption{Our framework DDD optimizes adversarial perturbations for images that protect malicious users from editing the images without consent. Through various test images, our efforts demonstrate that our approach is sufficient to cover copyrighted images, pornographic abuse, and public figure editing scenarios. Ultimately, it digresses the representation of the context image away from its multimodal nucleus in expectation.\\}
\label{page_1}
\vspace{-8pt}
\end{figure}

%% file: contents/1_introduction.tex
\section{Introduction}

The malicious editing of visual content is a long-standing ethical issue in the online community. Lately, this concern has only been significantly amplified with the integration of deep learning algorithms into the online community. Today, with recent advancements in the deep learning community, high-quality content crafting, known as Deepfake~\cite{sok,qad,bz,tw,all,why}, has been refined, and the discernment against real content is harder than ever. In tandem with malicious editing, this technology has led to increasing cases of social chaos and misinformation.

Recently, much interest has been directed to diffusion-based generative models~\cite{ho2020denoising,song2020score}. One impactful model among them is the text-to-image generator called Stable Diffusion~\cite{rombach2022high}, which was trained on LAION, a large-scale captioned image dataset~\cite{schuhmann2022laion}. This type of magnitude of training and model size unlocked the generative power of “foundational” scale and generalizability to unseen and complex prompts. With further fine-tuning \cite{zhang2023adding,hu2021lora,wang2023context,lin2023regeneration,xu2023prompt}, powerful variants such as inpainting models came to being, which allow the user to input a context image and inpaint the remainder with text guidance, quickly becoming the commercial inpainting approach. Additionally, a major advantage of inpainting over image editing algorithms \cite{mokady2023null,wallace2023edict} is that edits are exclusive to the user-defined region.

However, this created a breach for adversaries, as these models are well-suited for malicious and unconsented image edits of personal and private figures, copyrighted content, etc. This poses a serious problem on the internet, since the weights of foundational generative models are public to all individuals \cite{von-platen-etal-2022-diffusers}. To counter the production of such malicious content including deepfakes, researchers~ \cite{salman2023raising,ruiz2020disrupting} have proposed ways to disrupt their generation by injecting adversarial noise into the image to disrupt or fool either the face synthesizer or other subnetworks necessary in the generation. While considerable advancement has been made to disrupt GAN-based deepfake generators, the rise of diffusion-based deepfake pleads for an equivalent countering disruption algorithm. Unfortunately, because of the heterogeneous and complex characteristics underlying diffusion models, in which the generation process is gradual and iterative, previously proposed Deepfake disruption approaches are not compatible with the diffusion class of models. 

Photoguard~\cite{salman2023raising} has been proposed to address disruption for diffusion-based inpainters. It optimizes the context image so that it contains adversarial noise to disrupt the inpainting during inference time. While the effort of addressing the complications introduced by the diffusion process is noteworthy, it undergoes a significant computational overhead, amounting to 25GBs of VRAM and 19 min. of optimization. This computation cost is beyond the average consumer's budget. Furthermore, the disruption efficacy across different images, inpainting strengths, prompts is unstable.

Our work departs from the motivation that, while it is true that diffusion-based models introduce new technical challenges, we challenge the base assumptions that Photoguard considers to be necessary. In particular, to bypass the multiple feedforwards across the entire diffusion reverse process, we take into account the timestep range that, when perturbed with some adversarial signal, can cause the most amount of visual disparity, and harness the timestep constraint-free hidden states for our loss function. 

Next, we leverage the untargeted attack approach in the generative setting. Specifically, we find a centroid of representations to digress away from, which effectively eliminates the burden of defining a target point and its optimization complications of a targeted attack. First, we search for hidden state samples centered around the representation of the context image and calibrated through a discretely projected optimization in the token space. This optimized text embedding cooperates in defining a representative centroid through Monte Carlo sampling. This semantic-aware centroid faithfully captures the user's input in all modalities. Ultimately, distancing away from this point results in an edit that completely disassociates from the context image. Our “Semantic \textbf{D}igression-guided \textbf{D}iffusion \textbf{D}isruption”, DDD in short, is 3$\times$ faster, requires less GBs of VRAM, and the disruption is more effective than the current SoTA, Photoguard, across various images. We provide extensive results and experiments to support the integrity of our framework. Our main contribution is summarized as follows: 

\begin{itemize}
 \item In this work, we tackle disrupting inpainting-based deepfake synthesis with context optimization. We identify the most vulnerable timestep with respect to the feature space in the diffusion reverse process. This finding paves a design for a timestep-agnostic loss function, effectively decoupling our framework from the diffusion process' time/memory complexity overhead.

 \item We align our optimization objective with semantic digression optimization. Formally, we approximate the notion of synthesis correctness with a semantic-aware hidden state centroid, calibrated by a discretely projected optimization in the token space, and digress semantically away from it.
 
 \item The proposed framework, DDD, significantly reduces GPU VRAM usage and running time while sustaining effective disruption levels. Our effort democratizes image protection from malicious editing, bringing the computation expenses down to the consumer-grade GPU budget regime. 
\end{itemize}

%% file: contents/2_related_work.tex
\section{Related Works}

GAN-based deepfake generators~\cite{choi2018stargan,pumarola2018ganimation} have been the major consideration in both the research community and industry due to their long legacy, fast sampling, and quality. One distinguished architecture is StarGAN \cite{choi2018stargan}, trained to transfer the domain of the input image to cross-domains. And, GANimation~\cite{pumarola2018ganimation}, a conditional generator, can generate faces according to the expression annotation. 

To combat unconsented real image edits, Yeh et al.~\cite{yeh2020disrupting} pioneered the first attack on deep generative models, such as CycleGAN~\cite{zhu2017unpaired} and pix2pix~\cite{isola2017image}. They also introduced the Nullifying Attack to invalidate the model's generation and the Distorting Attack to make the generated images blurry or distorted. Ruiz et al.~\cite{ruiz2020disrupting} synthesize adversarial noise to the input image of these image-to-image GANs, so that their outputs are disrupted. Meanwhile, Wang et al.~\cite{wang2022deepfake} designed a perturbation generator, ensuring that the disrupted image output is correctly classified as a fake image by deepfake detectors. Furthermore, Huang et al.~\cite{huang2021initiative} introduced an iteratively trained surrogate model to provide feedback to the perturbation generator, enabling the suppression of facial 
manipulation.
While these disrupters are effective, the expressivity of the deepfake images made with GAN-based models is limited by the annotation, quality of the dataset, and model scalability. 

On the other hand, the recent text-to-image latent diffusion such as Stable Diffusion is trained on LAION~\cite{schuhmann2022laion}, a large-scale dataset scraped from the web. Naturally, it can generalize to complex and unseen prompts, and synthesize with high coherence. In particular, the Stable Diffusion (SD) Inpainter model~\cite{rombach2022high} is a finetuned Stable diffusion with masked context image conditioning and it can edit via inpainting the masked area. To disrupt such a potential deepfake generator, Photoguard takes a similar approach to previous disrupting algorithms in optimizing the adversarial noise in the context image. Their contribution and details will be provided in the next section.

%% file: contents/2_1Background.tex
\section{Background}

\subsection{Adversarial Attacks}
An adversarial attack is a method to generate adversarial examples to deceive the ML system~\cite{goodfellow2014explaining}. In the perspective of discriminators, given an objective function $\mathcal{L}$, a projective function $\Pi$, and an image $x$ its  true label $y$, the PGD~\cite{madry2017towards} attack is performed as follows:
\begin{equation}
x^{t+1} = \Pi_{x+\mathcal{S}}(x^t - \alpha ~ sign(\nabla_x \mathcal{L}(x,y_{target}))).
\end{equation}
This iterative update algorithm helps in identifying local maxima that induce misclassification and it is the engine behind most disruption frameworks. Likewise, we adopt PGD in our framework to update and refine our disruptive perturbation. 

\subsection{Diffusion Models}

Consider $x_t$ $\in \mathbb{R}^{1 \times 3 \times W \times H}$, where $x_T$ is an isotropic Gaussian noise, $x_0$ is true data observation, and any $x_t$ between these is defined as follows:
\begin{equation}
\begin{aligned}
q\left(x_t \mid x_{t-1}\right) & := \mathcal{N}\left(x_t ; \sqrt{\alpha_t} x_{t-1},\left(1-\alpha_t\right) \mathbf{I}\right) \\
q(x_T|x_0)&\approx\mathcal{N}(x_T;\mathbf{0},\mathbf{I}) \\
\end{aligned}
\end{equation}
Accordingly, diffusion models \cite{sohl2015deep,ho2020denoising,song2020score,song2020denoising} are a class of generative models that gradually denoises pure random noise $x_T$, until it becomes true data observation $x_0$. The training of such a model consists of predicting $x_{t-1}$ given $x_t$, where ground truth $x_{t-1}$ can be analytically yielded as an interpolation between $x_0$ and $x_t$ through Bayes rules. With reparametrization of $x_t$, it is rather common to predict the $\epsilon$ injected to $x_0$ to sample $x_t$, formulated as:

\begin{equation}
\mathcal{L}_\text{diffusion}=\mathbb{E}_{x_t, t, \epsilon \sim \mathcal{N}(0,1)}\left[\left\|\epsilon-\epsilon_\theta\left(x_t, t\right)\right\|_2^2\right]
\end{equation}

\subsection{Stable Diffusion Inpainters}
Following from the definition of diffusion models, the Stable Diffusion uses the same training paradigm \cite{rombach2022high} but diffuses in the latent space. Formally, $x_0$ is first encoded with a VAE encoder to the latent space as $z_0$ $\in \mathbb{R}^{1 \times 4 \times 64 \times 64}$, and the diffusion process occurs in the latent space, where $\epsilon$ is predicted, given $z_t$ and a text embedding condition $\tau$. These components change the formulation to:
\begin{equation}
\mathcal{L}_\text{LDM}=\mathbb{E}_{ t, c, \epsilon \sim \mathcal{N}(0,1)}\left[\left\|\epsilon-\epsilon_\theta\left(z_t, \tau, t\right)\right\|_2^2\right],
\end{equation}
where $\tau$ is the text embedding. The so-called “Inpainters'' are fine-tuned models from the Stable Diffusion checkpoints \cite{rombach2022high,zhang2023adding}. Typically, given some latent code $z_0$ of a real image $X$ to be inpainted, inpainting models minimally receive as input a context image latent $C$   $\in \mathbb{R}^{1 \times 4 \times 64 \times 64}$ and a binary mask $M$ $\in \mathbb{R}^{1 \times 4 \times 64 \times 64}$  as follows:
\begin{equation}
C = M \circ X,
\end{equation}
\begin{center}
$\text{where}\hspace{0.7cm} M_{i,j} = \begin{cases}
    0, &\small\text{if to be inpainted}\\
    1, &\small\text{if to be context}
\end{cases}$
\end{center}
In the Inpainter’s finetuning process, Stable Diffusion model's original input channels are extended so that both $M$ and $C$ can be fed as additional conditional signal for denoising $z_t$. Then, $M$ and $C$ will be conditioned throughout the iterative denoising of $z_T$ up to $z_0$.

In particular, the generation process spans over $n$ feedforward iterations, where $n$ commonly ranges between 20 to 50 for a reasonable generation. The backpropagation of $n$ feedforward is memory-wise infeasible. The current SoTA framework Photoguard~\cite{salman2023raising} optimizes the context image $C$ with PGD, the subject of protection from malicious edits. The authors of Photoguard approximate $z_0$ by iterating over just 4 denoising steps, which is assumed to be sufficient to synthesize an approximation $\hat{z_0}$. With this sample $\hat{z_0}$, $z_0$ is decoded to $x_0$ and the $L_2$ distance between $x_0$ and an arbitrary target image is minimized. Formally, with some simplification, Photoguard's loss is as follows:
\begin{equation}
\mathcal{\delta} = \underset{\lVert \delta \lVert_2 \leq \ \epsilon}{\operatorname{argmin}} \
 \lVert f(x+\delta, C) - x_{target} \rVert_2^2,
\end{equation}
where $f$ denotes the entire LDM pipeline over $n$ feedforward steps, $C$ is the context image, and $\delta$ is the adversarial perturbation. To the best of our knowledge, Photoguard is the only algorithm that tackles disruption in diffusion-based deepfake generation inpainters.

\begin{figure*}[t!]
\centering
\includegraphics[width=\textwidth]{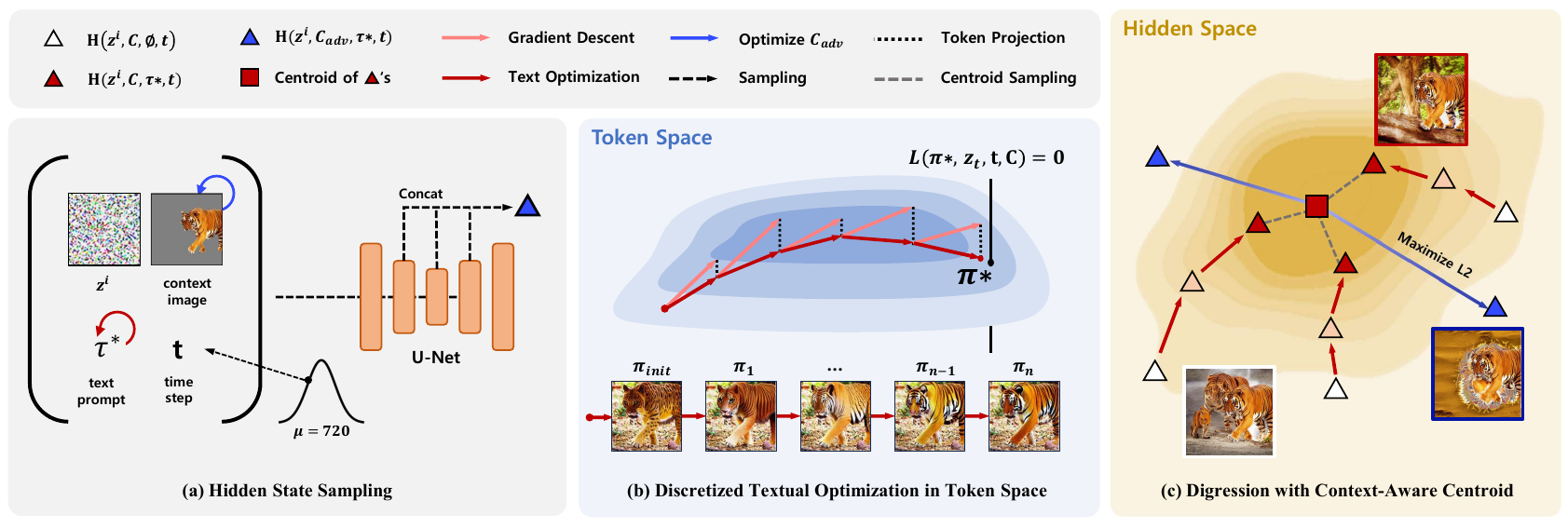}
\caption{Overview of DDD's Framework: Our framework's objective lies in finding the context image's representative multi-modal centroid, for which our immunized image's representation semantically digresses away from it. (a) illustrates the pipeline for sampling all of the hidden states utilized in the framework. In (b), we first utilize our context image to yield the diffusion-based inpainting loss to update the token embedding $\pi^\ast$.  Finally, with $\tau^\ast = \mathcal{E}(\pi^\ast)$, we construct the multi-modal centroid via Monte Carlo sampling.}
\label{fig:Framework}
\vspace{-8pt}
\end{figure*}

%% file: contents/3_method.tex
\section{Method} \label{methodology}
\subsection{Search for the Vulnerable Timestep}

One of the most self-evident complications presented by SD Inpainter is the diffusion process. To take into perspective, the denoising diffusion reverse process can be thought of as the random latent code being decoded by a VAE~\cite{oord2017neural}. Then, it is plausible that we need to feedforward through all synthesis “stages” to optimize efficiently. 

Instead of directly complying with this computational overhead, 
we take advantage of the ``progressive synthesis" property of the diffusion process to strategically target the early timesteps.
It is known to the community that early timesteps have sovereignty over the overall spatial structure and global semantics of the image \cite{meng2021sdedit,huang2023reversion,chen2023beyond}. 
To gain a sufficient degree of freedom of disruption, we optimize our adversarial context with respect to the early stage, leading to global damage. It is noted that the model should behave similarly across timesteps adjacent to our target timestep range to sustain the vulnerability of predicted scores $\nabla_x \log p(x)$.  While many researchers already rely on the linearization of adjacent timesteps ~\cite{huberman2023edit,miyake2023negative}, we reaffirm this linearity through both PCA decomposition of the hidden states across all timesteps, discussed in Appendix \ref{appendix:Timestep}, and empirical results of effective disruption.

\subsection{Timestep Constraint-Free Loss Function}

While the narrowing of the sampled timestep disables access to both the diffusion-native and off-the-shelf losses (LPIPS, CLIP, Perceptual loss), we take advantage of the denoiser's hidden states to yield a timestep-agnostic loss space. Specifically, we are concerned with strategically drifting the hidden state out of its learned manifold in the Euclidean space. Then, if we have the text embedding $\tau_{text}$ obtained from the text encoder $\mathcal{E}$ and the feature map $F_{C,\tau_{text}}$, our loss on the space of $H$ is as follows:
\begin{equation}
\mathcal{L}_{hidden} = \sum_{i}^{N} \lVert H^{target}_i - H^{source}_i \rVert^2_2,
\end{equation}
\begin{center}
$\text{where}\hspace{0.7cm} H(C,\tau_{text},z_T) = \text{Attention}(F_{C,\tau_{text},z_T}) $
\end{center}
\noindent $F_{C,\tau_{text},z_T}$ is the feature map conditioned with $C$, $\tau_{text}$. While all three conditions are indispensable for $F$, we omit their inscription when they are constant. Because we want to disrupt the information pathway for multi-modal signals, ($C$ and $\tau$),  we choose the self-attention layers. Ultimately, this loss space enables the mining of both dense and timestep-agnostic features.  

\begin{figure*}[ht]
 \centering
  \includegraphics[width=\textwidth]{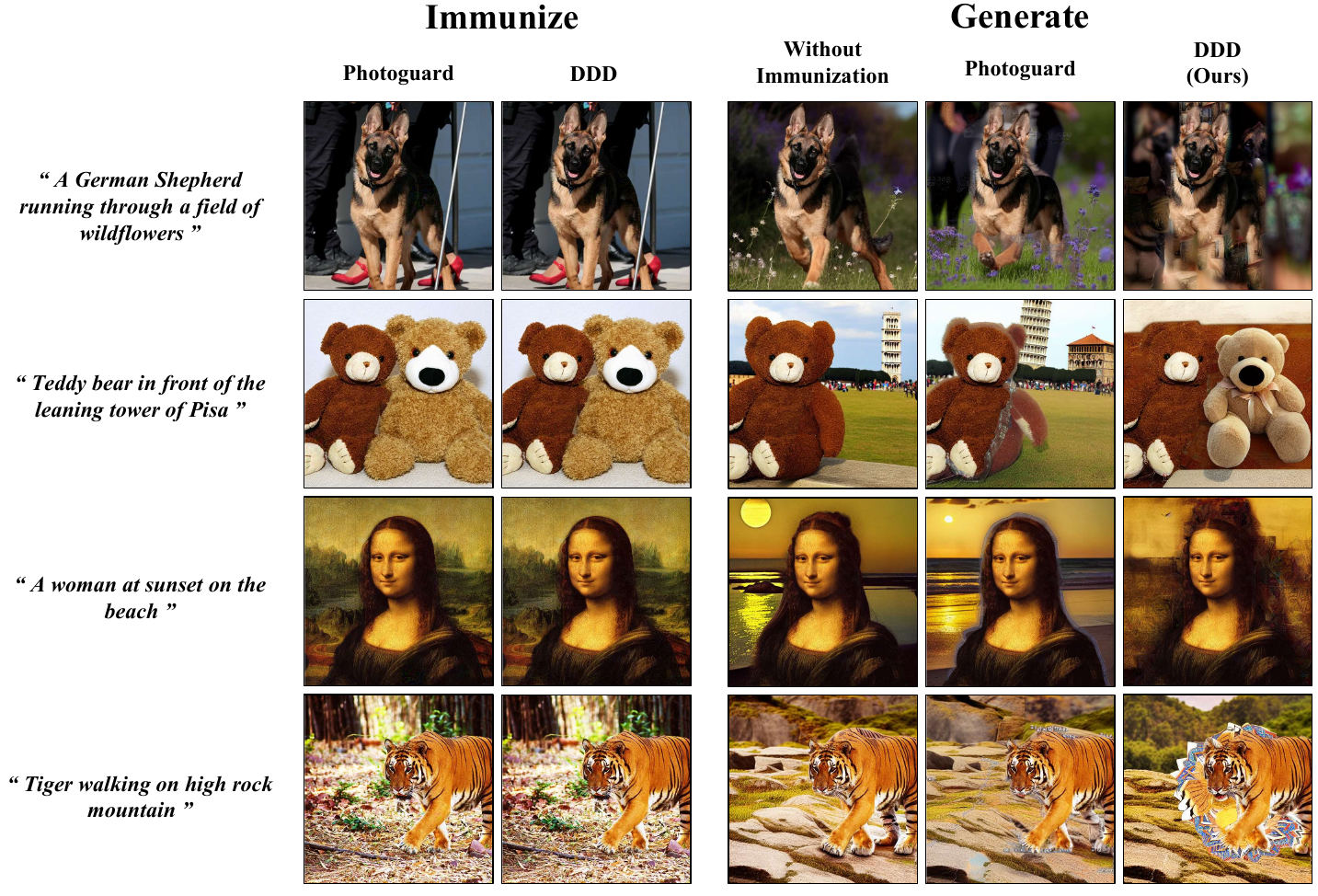}
 \caption{A side-by-side comparison of Photoguard against DDD. DDD disrupts the hidden representations, which leaves more clearly visible unnatural artifacts and disruption across the images.}
 \label{fig:qualitative}
 \vspace{-8pt}
\end{figure*}

\subsection{Aligning the Objective with Disruption's Goal}

The success of untargeted attacks is well-known to the adversarial attack community.

In our setting where the sole concern is pulling away from what is correct, the untargeted attack approach is highly aligned with our implicit objective, with no other constraints imposed other than disruption itself.

Now, it is noteworthy that in the discriminative setting, the notion of "correctness" is trivially given by the probability vector. Then, cross-entropy gives us a scalar metric to evaluate the correctness. However, for the most part in generative models, this type of reduced evaluative metric is not plainly given, hence $y_{true}$ is ambiguous. As an alternative tentative approach, we discuss the consequences of framing the approach with a targeted attack in Appendix \ref{appendix:Discussion_TO}.

To adopt untargeted attacks to diffusion models, we approximate the concept of a correct synthesis. A simple approach to obtain the correct output synthesis is to take the average at the $x_t$ space in $\mathbb{R}^{D_{pixel}}$, where $D_{pixel} =  W \times H$. Unfortunately, such averaging yields a low-fidelity and ambiguous ground truth target (equivalent to an averaged blurred image). Instead, the hidden space $\mathbb{R}^{D_{hidden}}$, where $D_{hidden} << D_{pixel}$, is suitable for yielding a high-fidelity centroid with less variance and higher spatial agnosticism. Accordingly, through Monte Carlo sampling, we approximate the ground truth representation centroid $\phi_{hidden}$, defined as following: 
\begin{equation}
\phi(C,\tau_{text}) =\mathbb{E}_{z_T \sim \mathcal{N} (0,~1), ~ t \sim \mathcal{N}(720,~5.8)}[H(z_T,~ C,~ \tau_{text}, ~t)]
\end{equation}

While the context image is given apriori, the choice of the text prompt $\tau$ is still ambiguous. In other words, given that the deep activations accommodate for multi-modal signal \cite{hussain2023information,elhage2021mathematical}, ignoring the text modality is prone to yield sub-optimal results.

\newcolumntype{g}{>{\columncolor{Gray}}c}
\begin{table*}[h]
    \centering

  \label{tab:metrics}
    \begin{tabular}{lccccccll}
    \toprule
     &CLIP Score $\downarrow$  &Aesthetic $\downarrow$ & PickScore$\downarrow$& FID $\uparrow$ &  KID $\uparrow$&LPIPS $\uparrow$&SSIM $\downarrow$&PSNR$\downarrow$\\
    \hline
    (R $\rightarrow$ R) Photoguard& 0.27819&5.5855& 0.69& 66.25& 0.00183&0.4262& 0.5817&29.6200\\
    \rowcolor[gray]{0.9}
    (R $\rightarrow$ R) DDD& \B 0.24429&\B5.2246&\B0.31&\B118.97&\B0.02093&\B 0.4993& \B0.5153&\B 29.3763\\
    
    \hline
    (R $\rightarrow$ S) Photoguard & 0.27842& 5.6501& 0.56& 65.83& 0.00157& 0.3600& 0.6521& 29.8143\\
    \rowcolor[gray]{0.9}
    (R $\rightarrow$ S) DDD 
    & \B0.27388& \B5.4194& \B0.44& \B80.54& \B0.00481&\B0.4032&\B0.6129&\B29.6729\\
    
    \hline
    (S $\rightarrow$ S) Photoguard & 0.27344&5.5592& 0.63& 70.98& 0.00315& 0.3928& 0.6338& 29.6738\\
    \rowcolor[gray]{0.9}
    (S $\rightarrow$ S) DDD & \B0.25074& \B5.2805&\B 0.37& \B106.13& \B0.01345&\B 0.4423& \B 0.5751&\B 29.5184\\
    
    \hline
    (S $\rightarrow$ R) Photoguard & 0.27984&5.5605& 0.52& 63.89& 0.00200&0.4011& 0.6033&29.6722\\
    \rowcolor[gray]{0.9}
    (S $\rightarrow$ R) DDD &\B 0.27412&\B 5.5284& \B0.48& \B72.62& \B0.00312& \B0.4084& \B0.5859& \B29.6683\\
    
    \hline
    Oracle (R)&0.2582  &5.8147&-&-&-&-& -&-\\
    Oracle (S)&0.2597  &5.8168&-&-&-&-& -&-\\
   \bottomrule
    \end{tabular}
    \caption{Quantitative Metrics of Disruption. We also show results for the disruption in the transferability setting. ``R'' stands for Runway 1.5v and ``S'' for Stability AI 2.0v, where R $\rightarrow$ S means disruption optimized with R and applied on inpainting with S.  The arrows alongside the metric names show the desirable direction from the perspective of disruption. 
    }
\end{table*}

\subsection{Token Projective Textual Optimization}

One simple approach is to condition over the null text, which yields a reasonable centroid that captures the signal of the context image. However, the null text is also implicitly biased and fitted to the global dataset, which is naturally long-tailed. To this end, we address this by searching in the text embedding space for the representation that embodies the context image's content and use it to condition upon our centroid-constituting instances. Specifically, we optimize text embedding $\tau$ that minimizes the diffusion loss:
\begin{equation}
\tau^{\ast} = \underset{\tau}{argmin}~ || \epsilon * M - \epsilon(z_T, C, \tau, M^{\text{inv}}) * M ||_2^2,
\end{equation}

\noindent where $M^{\text{inv}}$ is $M$ inverted and $\tau$ is initialized from a random text embedding. One common issue with most image inversion algorithms ~\cite{zhu2020domain,xia2022gan} is the propensity of the embedding to overfit to the reference image, which results in anomalous expressivity and poor generalization. While premeditatively underfitting it viable, convergence points vary for every instance.

PEZ has shown that one can invert a reference image with ``decodable" token embeddings ~\cite{wen2023hard}. Formally, in between every gradient descent, the updated token embeddings are projected to their nearest tokens, which are ultimately used for text decoding. Instead, we view this intermediate discretization as an approach to optimization regularization. In essence, the token embedding matrix constitutes the keypoints for the topology of the token space manifold. Because this token embedding matrix encompasses the primal semantics, having the continuous embedding projected to this learned manifold ensures that it stays in-distribution.
Then, while a constraint-free embedding optimization eventually leads to overfitted solutions, the inherent discreteness of the token projection intercepts any adversarial build-up.

Formally, let $\pi$ be token embeddings, $Proj$ be the token projection function, $\mathcal{E}$ be the text encoder, $\alpha$ be the step size, and $\mathcal{L}$ be the diffusion loss, we optimize for the token embedding that minimizes the diffusion loss with a projective update as follows:
\begin{equation}
\pi_{t+1} = \pi_t - \alpha \nabla_{\pi_t} \mathcal{L} (\mathcal{E}(Proj(\pi_t)), z_t, t, C)
\end{equation}

Having obtained the optimized $\pi^\ast$, we have implicitly yielded its bijective counterpart $\tau^\ast = \mathcal{E}(\pi^\ast)$, the text embedding. Ultimately, this optimization will lead to a $\tau^\ast$ highly descriptive of the context image, and the centroid $\phi(C, \tau^{\ast})$ will embody the expected multi-modal representation of the context image. Given that we obtained this surrogate for $y_{true}$, we now have access to the digression formulation in MSE, which, due to what the centroid is encoding, the optimization will lead to a digression orthogonal to the semantics of the context image. Ultimately, our final loss is as follows:

\begin{equation}
C^{\ast} = \underset{C}{argmax}~|| 	\underbrace{\phi(C,\tau^{\ast})}_{\text{target}} - \underbrace{H(z_T, C, \tau^{\ast}, M)}_{\text{source}}||_2^2
\end{equation}

The iterative update of $C$ follows the PGD update protocol. Ultimately, as summarized in Fig.~\ref{fig:Framework}, due to its digressive objective, we name our framework ``Digressive Guidance for Disrupting Diffusion-based Inpainters", or in short, DDD. 

%% file: contents/4_experiment.tex

\section{Experimental Results}

\subsection{Technical Details}
For DDD's disruption synthesis, we have used Nvidia's A100 40GB GPU, taking under 6 minutes of optimization, more than 3 times faster than Photoguard. All of our experiments are conducted with PGD's epsilon budget of 12/255, step size of 3/255, and gradient averaging of 7 for 250 iterations. While the loss curve hints that further optimization is possible with noticeable returns, we verified that 250 is enough to cover all examples. Additionally, our maximum memory usage is 16 GBs.

\subsection{Quantitative Results}
Taking into consideration that inpainting disruption in diffusion models is an emerging task, we present a variety of metrics to explore different angles of analysis of each disruption method. For all of these metrics, we curated a dataset with in-the-wild and heterogeneous images, totaling 381 pairs of images and prompts. We test over different inpainting strengths (0.8, 0.9, 1.0) to verify their disruption robustness.

The quantitative metrics can be sub-categorized into two groups. 
First, metrics such as CLIP Score, (CLIP) Aesthetic, and Pick-a-Score evaluate the alignment of the inpainted image and its prompt. CLIP Aesthetic and Pick-a-Score are especially insightful due to their inherent design to rank images upon their visual integrity. In this category, DDD achieves a perfect score against Photoguard. The remaining metrics, SSIM, PSNR, FID, and KID, evaluate the visual disparity between the oracle inpainting and the disrupted inpainting through deep and natural metrics. Similarly, DDD outperforms Photoguard.

\begin{figure}[h]
\centering
\includegraphics[width=\columnwidth]{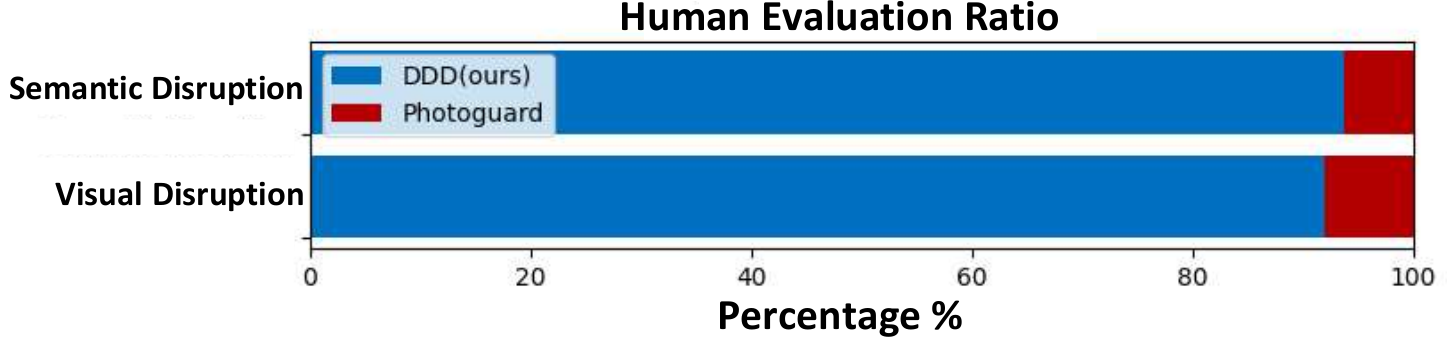}
\caption{Human evaluation on 20 random disrupted examples.}
\label{fig:human_eval}
\vspace{-8pt}
\end{figure}

\begin{figure*}[h]
\centering
\includegraphics[width=\textwidth]{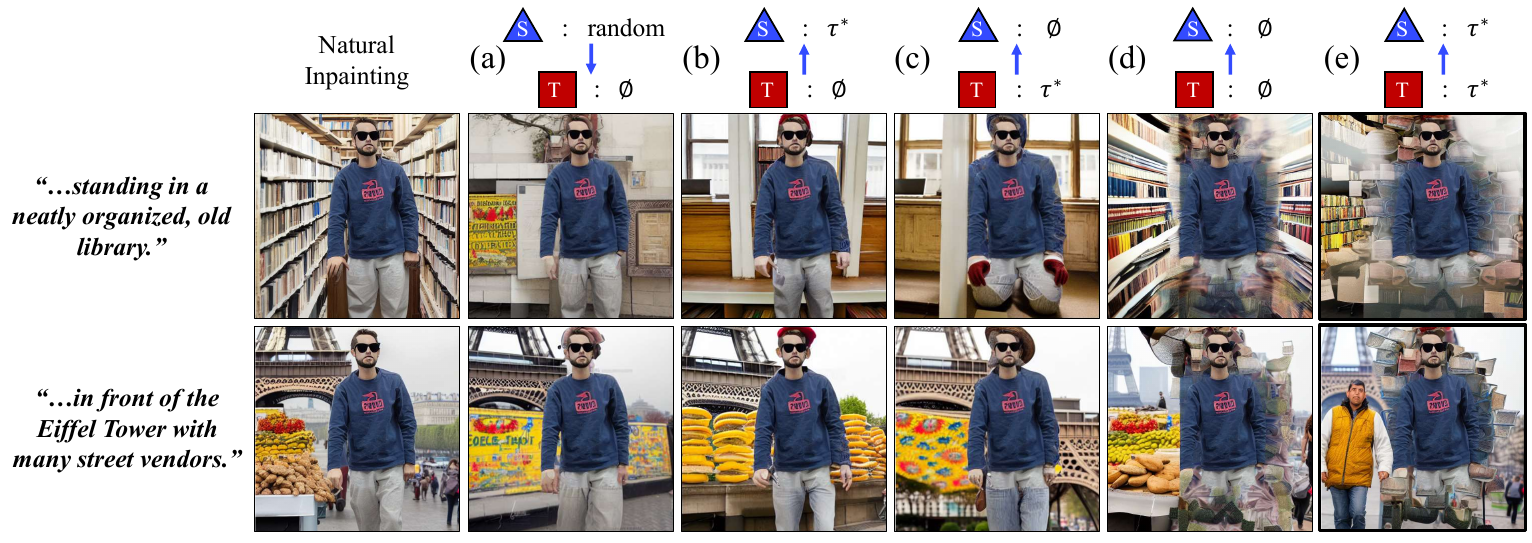}
\caption{An ablation study for various targeted and untargeted scenarios. In this context, ``S'' and ``T'' represent the text conditions of source and target, respectively. Furthermore, the down arrow ($\color{blue}\downarrow$) signifies a targeted scenario, where the objective is to move closer to the fixed $H^{target}$, while the up arrow ($\color{blue}\uparrow$) represents an untargeted scenario, where the objective is to move away from the fixed $H^{target}$. And, (e) represents the result from our work, DDD.}
\label{fig:ablation}
\vspace{-8pt}
\end{figure*}

\subsection{Qualitative Results}
The qualitative evaluation rules over the success of inpainting disruption. In Fig.~\ref{fig:qualitative}, we present images, create their respective masks, and test the inpainting under no disruption, Photoguard, and our method DDD. Photoguard's disruption disassociates the context image's spatial attributes from the inpainted area. However, a large portion of the disrupted images show weak or failed disruption. DDD, on the other hand, shows consistent disruption.


As this task directly involves end users, their evaluation is indispensable. Hence, we present results on human evaluation from 31 human evaluators of Photoguard and DDD-based disruptions. We randomly chose 20 image disruption examples and collected responses on the textual and visual criteria. As shown in Fig.~\ref{fig:human_eval}, 91.94 \% and 93.85 \% of the evaluators found Photoguard to display higher semantic alignment and visual coherence, respectively, with 0.1 and 0.06 standard deviations. These results imply that DDD's disruptions are more closely aligned with the basic disruption criteria. More details on the experimental setup are shared in Appendix \ref{appendix:humaneval}. 

One subtle detail in these two competing frameworks lies in their sensitivity to different inpainting strengths, as shown in Appendix \ref{appendix:Inpainting_st}. Although easily glanced over, the discussion of disruption under different strength thresholds is significant. Since we do not know the specific type of edit the end-user will execute apriori, Photoguard and DDD should disrupt the best when strength is close to 1.0 since no image signal is encoded at $z_T$ and the disruption has more room for digression. At strength = 0.8, Photoguard shows weak disruption across some images while our method shows consistent and effective disruption.

\subsection{Ablation Study}
The right construction of centroid $\phi$ is crucial for asserting disruption to unseen context images and prompts. Put differently, if $\phi$ is at the nucleus of the context image's representation, or the expected representation, deviating away from it encourages orthogonal semantic digression. To bring light to the representative effect, we exhibit five pairs of sources and targets to compare with DDD, each with different objectives used during optimization. In Fig.~\ref{fig:ablation}, 
each setting deals with an objective where the distance between hidden representations of the source $H^{source}$ and target $H^{target}$ are either minimized (targeted) or maximized (untargeted). Specifically, $H^{random}$ refers to a set of hidden states \{$H(C,\tau_{0})$, $...~, H(C,\tau_{j-1}), H(C,\tau_{j})$\}, where $\tau_j$ is a collection of $j$ text embeddings of randomly chosen prompts to express an i.i.d. distribution of text.

\textbf{Targeted Scenario.} In Fig.~\ref{fig:ablation}(a), mild disruption is achieved when $H^{random}$ is pulled to $\phi^{null}$. While the inpainting deviates from the text condition, rich semantics such as Eiffel remains intact. While this setting can be considered as an alternative approach depending on the specific user preference, the optimization resources, namely the number of tunable pixels of the context image, are limited and under-expressive to match a distribution of hidden states to a single point. This leads to image protection failures in a significant portion of the test cases.

\textbf{Untargeted Scenario.}  Given that $\tau^\ast$ is the text that fully represents the context image's content, our method follows (e) in Fig.~\ref{fig:ablation}, where the semantic digression of $H(\tau^\ast)$'s are heading directly away from $\phi(\tau^\ast)$. As an experimental cause, we show (b) and (c), where we premeditatedly construct the digression to lose its orthogonality with respect to the $C$'s semantics. Additionally, thanks to our framework's formulation, (d) shows competitive performance. However, (e) outperforms in all metrics. Namely, we have calculated the averaged scores across different checkpoints and in the transferability setting and obtained superiority in every metric. We append the full results in the Appendix Table \ref{tab:metrics_ablation}.

%% file: contents/5_Discussion.tex
\section{Discussion}
Our work entails discussions from various ends. Particularly, note that DDD is a conceptual framework. Put differently, the applicability of our framework is compatible with untargeted disruption of upcoming variants of inpainters and synthesizers, even generalizing beyond diffusion models.

Inpainting disruption is an emerging task, especially with the introduction of powerful text-to-image models. The current literature on inpainting disruption is limited and has much room for growth. For example, no disruption-specific metric has been proposed yet. As of now, human evaluation and PickScore are the metrics with the highest fidelity in grading the inpainting synthesis completeness. 

Also, we found that both DDD and Photoguard need more robustness to data augmentations and agnosticity to the context image's mask positioning. We believe this field will highly benefit from these developments. In addition to these topics, we discuss experimental configurations of our framework, strength interplay, transferability, survey form, metrics, and more in the Appendix.

%% file: contents/6_conclusion.tex
\section{Conclusion}
We proposed to immunize context images and disrupt their malicious inpainting edits through our framework DDD. To strategically manage the diffusion process, we focused on the model's hidden state statistics across different timesteps. This showed us how to select a vulnerable timestep range and reduced the complexity of the attack's forward pass. 
Next, we formulate the disruption objective as a semantic digression optimization, which not only offers a greater degree of optimization, but also takes full consideration of the context image. This is only possible through searching for the multi-modal centroid, calibrated and regulated by token projective embedding optimization, and constructed through Monte Carlo sampling. As a result, we significantly reduce GPU VRAM usage and speed up the optimization time by 3$\times$. 
Our framework is supported by quantitative and qualitative results, and contributional research resources on this novel task in diffusion inpainters.

\section*{Ethical Statements}
Our work involves nudity and sexually explicit content, but as all models are publicly available, our institution's IRB advised that approval was not required. All researchers involved are over 21 and have carefully reviewed relevant ethics guidelines~\cite{g1} and undergone training to handle and analyze research results properly. Although no practical defense against creating nudity in generative models exists, we emphasize the urgency of developing preventive technologies given our work's focus on explicit and unsafe content.

\section*{Acknowledgments}
The authors would thank anonymous reviewers. Geonho Son and Juhun Lee contributed equally as joint first authors. Simon S. Woo is the corresponding author. This work was partly supported by Institute for Information \& communication Technology Planning \& evaluation (IITP) grants funded by the Korean government MSIT: (No. RS-2022-II221199, Graduate School of Convergence Security at Sungkyunkwan University), 
(No. RS-2024-00337703, Development of satellite security vulnerability detection techniques using AI and specification-based automation tools), (No. 2022-0-01045, Self-directed Multi-Modal Intelligence for solving unknown, open domain problems), (No. RS-2022-II220688, AI Platform to Fully Adapt and Reflect Privacy-Policy Changes), (No. RS-2021-II212068, Artificial Intelligence Innovation Hub), (No. 2019-0-00421, AI Graduate School Support Program at Sungkyunkwan University), and (No. RS-2023-00230337, Advanced and Proactive AI Platform Research and Development Against Malicious Deepfakes), and (No. RS-2024-00356293, AI-Generated Fake Multimedia Detection, Erasing, and Machine Unlearning Research).

%% file: contents/appendix.tex
\section*{Appendix}

\begin{figure}[h]
 \centering
  \includegraphics[width=\columnwidth]{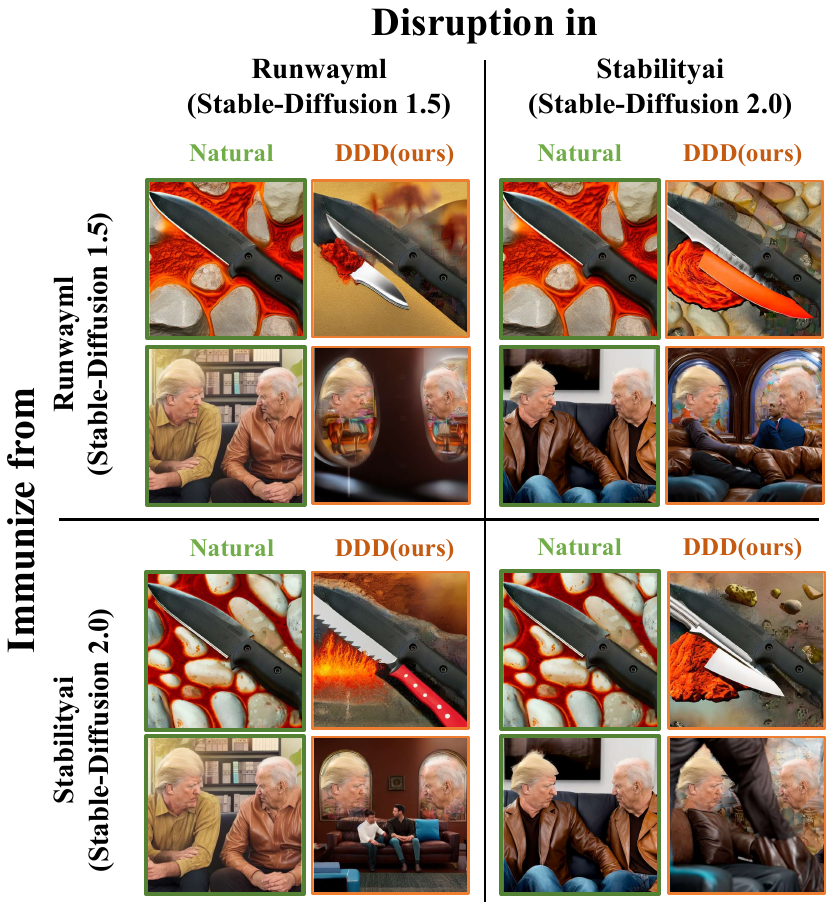}
 \caption{Transferability and framework generalization across different model versions. This figure allows for a comparison of how much disruption occurs from different models after the image has been immunized by DDD. The prompts are "A knife floating on boiling red lava and stones" and "Two men sitting on a sofa [...] brown leather sofa" for upper and lower image, respectively. }
 \label{fig:transferability}
 \vspace{-8pt}
\end{figure}

\section{Immunization Transferability}\label{appendix:Immunization_trans}
Transferability refers to the optimization of the adversarial context image on one network to only be applied on a different network. To the extent of open-sourced foundational inpainter checkpoints, we check transferability between Runwayml's 1.5v model and Stabilityai's 2.0v checkpoint, the main checkpoints in the community. It is noteworthy that while these two share the same network architecture, they are independent checkpoints with different training protocols. In addition to the quantitative results in Table~\ref{tab:metrics_ablation}, Fig.~\ref{fig:transferability} qualitatively shows that our model transfers disruption bilaterally. On the other hand.

\begin{figure}[h]
 \centering
  \includegraphics[width=\columnwidth]{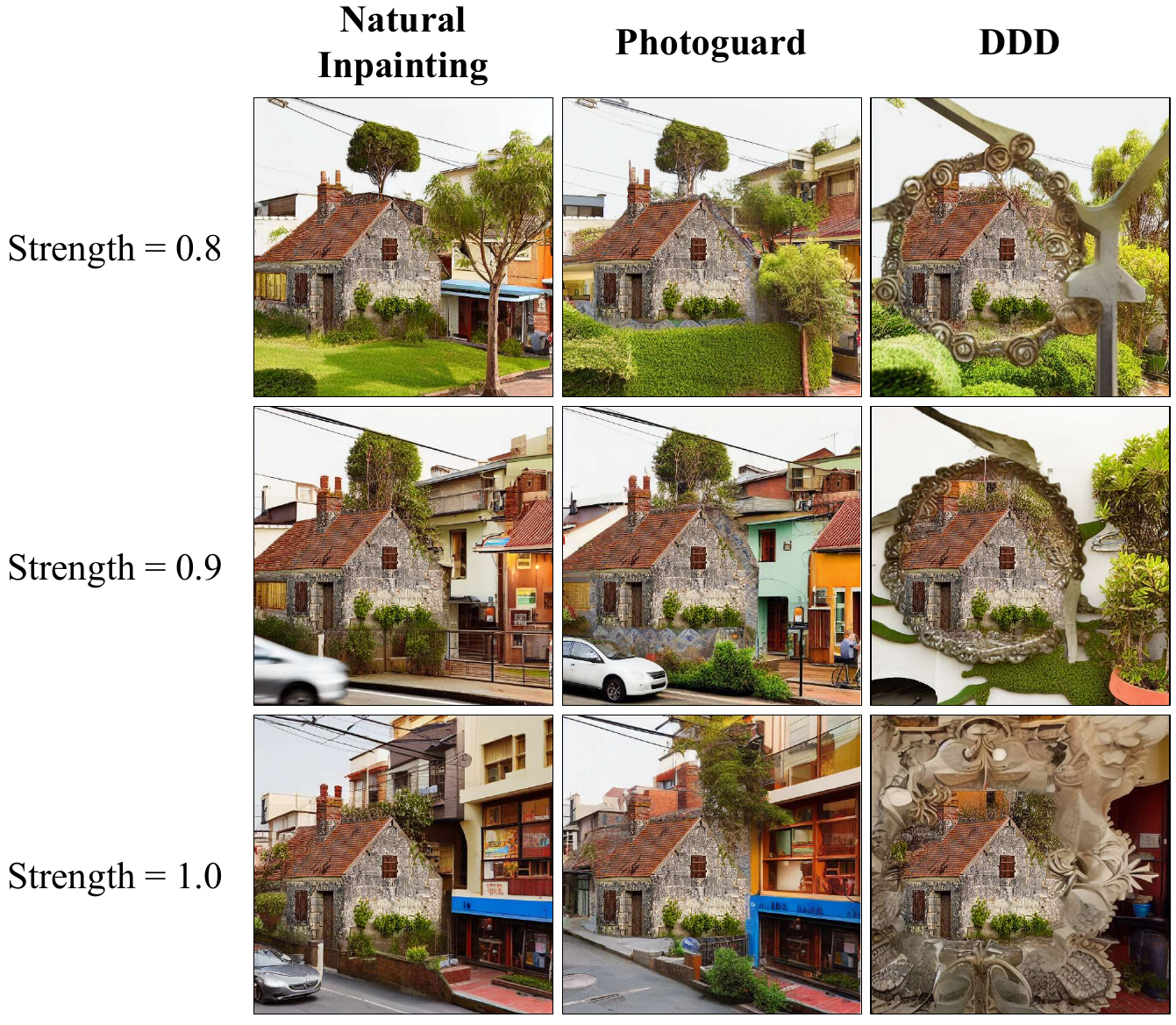}
 \caption{Strength sensitivity allows the user to control how much of the original image we will retain in the inpainting process. When there is no preservation with strength = 1, the inpainted image is exposed to the maximum disruption. The inpainting prompt used is ``A house on a street, surrounded by shops and cafes''.}
 \label{fig:strength}
 \vspace{-8pt}
\end{figure}

\section{Inpainting Strength}\label{appendix:Inpainting_st}
We also take note of the strength functionality of the SD Inpainter. Given a real image, strength decides how much content of this image we will preserve in the to-be-inpainted area. For example, considering the timestep $t$ to be normalized between 0 and 1, strength = 1 completely noise the inpainted area into $x_1$, whereas strength = 0.8 noises the area up to $x_{0.8}$. While crafting the adversarial noise by sampling close to $z_1$ ($z_T$) can prove to be the strongest for inpainting with strength=1, immunization considering all test case scenarios and cover harder thresholds such as 0.8. Therefore, we consider both the strength factor and the similarity between the hyperfeatures ~\cite{luo2023diffusion} to define our sampling timesteps range as $t \sim \mathcal{N}(720, 6)$. The sampling distribution is directly inferred from the previously discussed factors of considerations and shows robust performance even with an offset of 720 $\pm10$.

\section{Robustness}\label{appendix:Robustness}

\begin{figure}[h]
\centering
\includegraphics[width=\columnwidth]{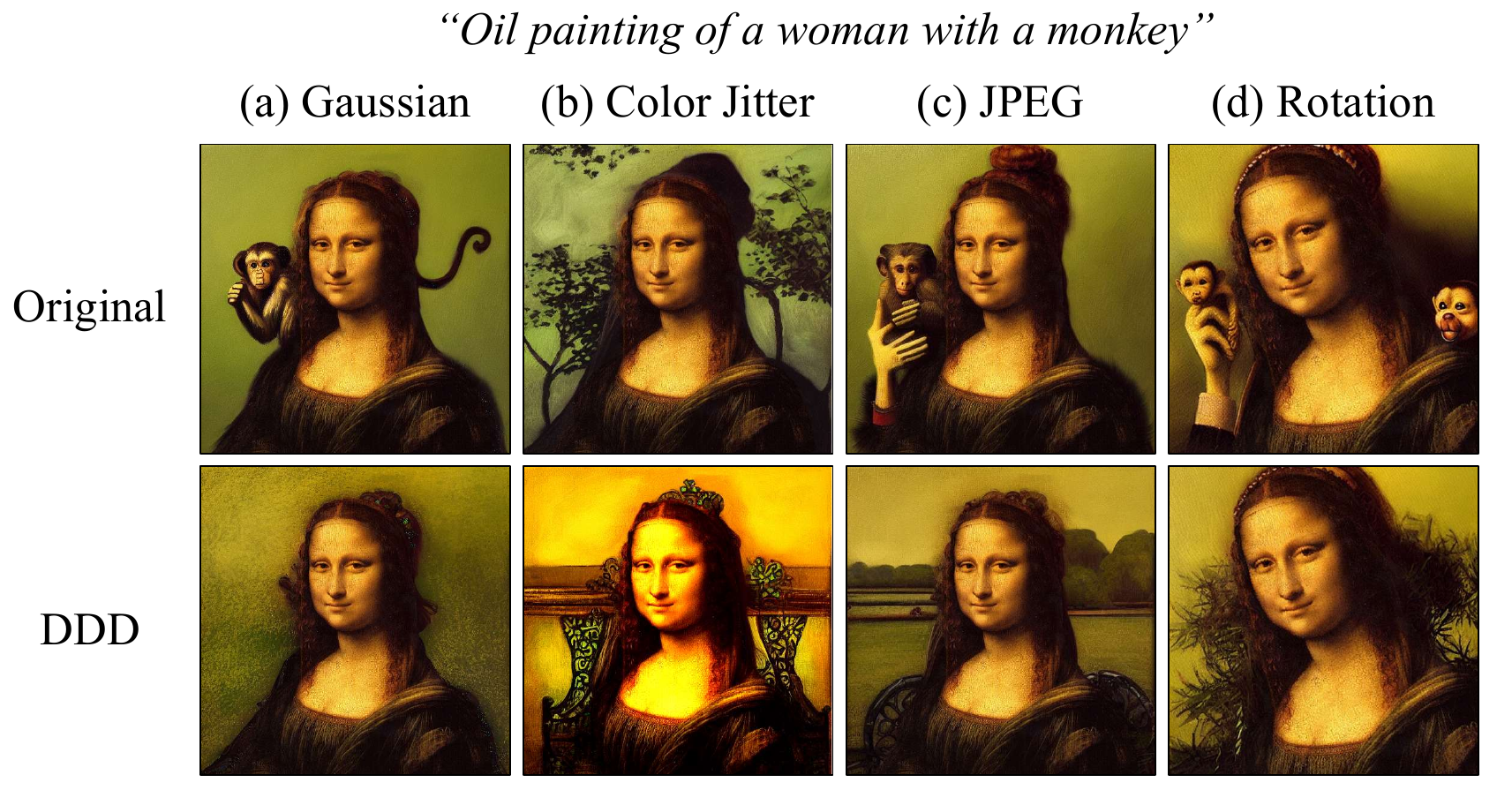}
\caption{Performance under different augmentations with DDD's highest tolerating thresholds.  
rotate (5 degrees) with resize cropping, Gaussian noise $N$(0,5), JPEG compression quality 80 (0 \text{\textasciitilde} 100), jitter (brightness 1.1 contrast 1.1 saturation 1.1). The first column shows inpainting performed to augmented images and the remaining columns show inpainting under respective augmentations.}
\label{fig:robustness}
\vspace{-8pt}
\end{figure}

Another important factor for image protection in the web is robustness to data augmentations and compression. As shown in  Fig.~\ref{fig:robustness}, we test if the adversarial perturbation survives across Gaussian noise, color jitter, JPEG compression, and rotation with crop. While the disruption performance degrades by a considerable margin, some images still retain the disruptive signal. Note that these are only representative cases where the adversarial signal survives different augmentations, but more research has to be conducted to improve its effect retention. Additionally, the task of disruption defines the protected area apriori. Obstruction of the adversarial noise is also expected to result in less effective attacks. Thus, these areas under robustness are of utmost contribution in this area in future works.

\section{Discussion on Targeted Optimization}\label{appendix:Discussion_TO}
The targeted optimization approach in both frameworks is how the disruption is bounded by when $\mathcal{L}_{targeted} = 0$. At this stage, it reaches the maximum disruption possible under the chosen $\mathcal{L}$ and some target image/latent. Yet, there is little guarantee that this maximum disruption will be sufficient and may fall short of our evaluation criteria. The success of the targeted approach relies on the search for an optimal target, which is an approach that asks for non-trivial considerations.

Here, Photoguard formulates a targeted attack supervised by the MSE loss with respect to an arbitrary target image. Accordingly, we attempted to take the targeted attack approach with our custom loss function and minimize the distance to some target, either conditioned on an arbitrary image, or text embedding. However, just like Photoguard, both methods do not show a high success rate and effective disruption. In fact, within our framework, we find that the attack’s success highly depends on highly tuned text prompt conditions for our source and target sample given a context image, which hints poor generalizability.

\section{Considerations on Diffusion Timesteps}\label{appendix:Timestep}
\begin{figure}[h]
\centering
\includegraphics[width=\columnwidth]{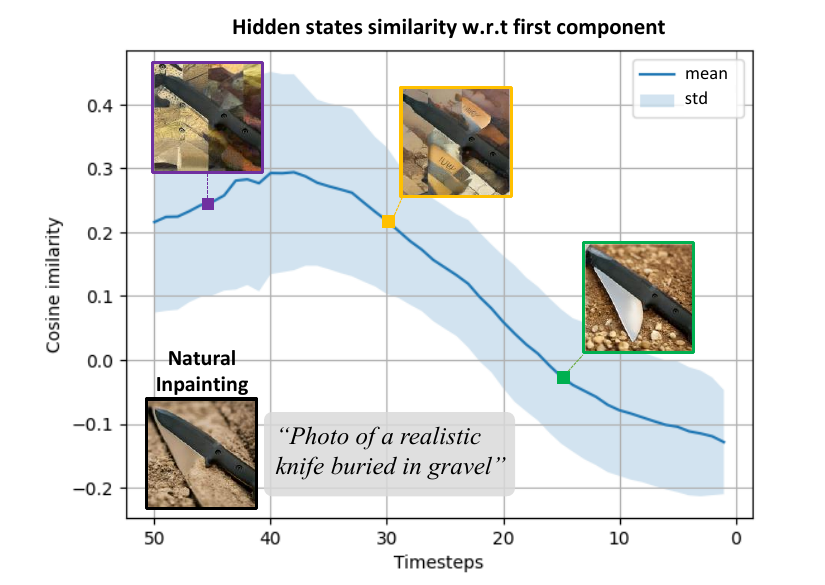}
\caption{PCAs of hidden states from resolutions 32$\times$32 and 8$\times$8 over all timesteps were calculated and averaged over three examples. The graph shows the similarity between PCA's first component and hidden states across all timesteps. Higher cosine similarity means higher homogeneity with respect to the timestep-wise Eigenfeature.}
\label{fig:pca}
\vspace{-8pt}
\end{figure}

One of the key designs for why DDD is faster than Photoguard lies in the non-iterative feedforwarding. This is motivated by the hypothesis that, while the chained feedforwarding is inevitable for image generation, one does not need the whole iterative chain. Treating the iterative diffusion process as a sequence of synthesis stages, we focus primarily on disrupting the initial stage, where most of the spatial structures are defined. However, this could only be effective if synthesizing an adversarial context image for a very particular timestep range has its disruption spilled over to sufficient adjacent timesteps. One of our findings that reassures the effectiveness of an adversarial context image in respect to adjacent timesteps is the level of similarity of the hidden states across different timesteps. We extract intermediate activations of different layers across all timesteps and apply principal component analysis (PCA) to uncover the ``Eigenfeature'', as shown in Fig.~\ref{fig:pca}, and calculate the cosine similarity between it and the hidden states from all timesteps. Note that a single global adversarial context image may not cover the whole breadth of timesteps. However, the early timesteps possess sufficient similarity, that when optimized with respect to them, the disruption effect will be shared across. Another benefit of targeting the early timesteps is that disruption at this level will directly affect the structural arrangement of the image.

\section{Token-Projective Optimization}\label{appendix:TPO}
\begin{figure*}[h]
\centering
\includegraphics[width=\textwidth]{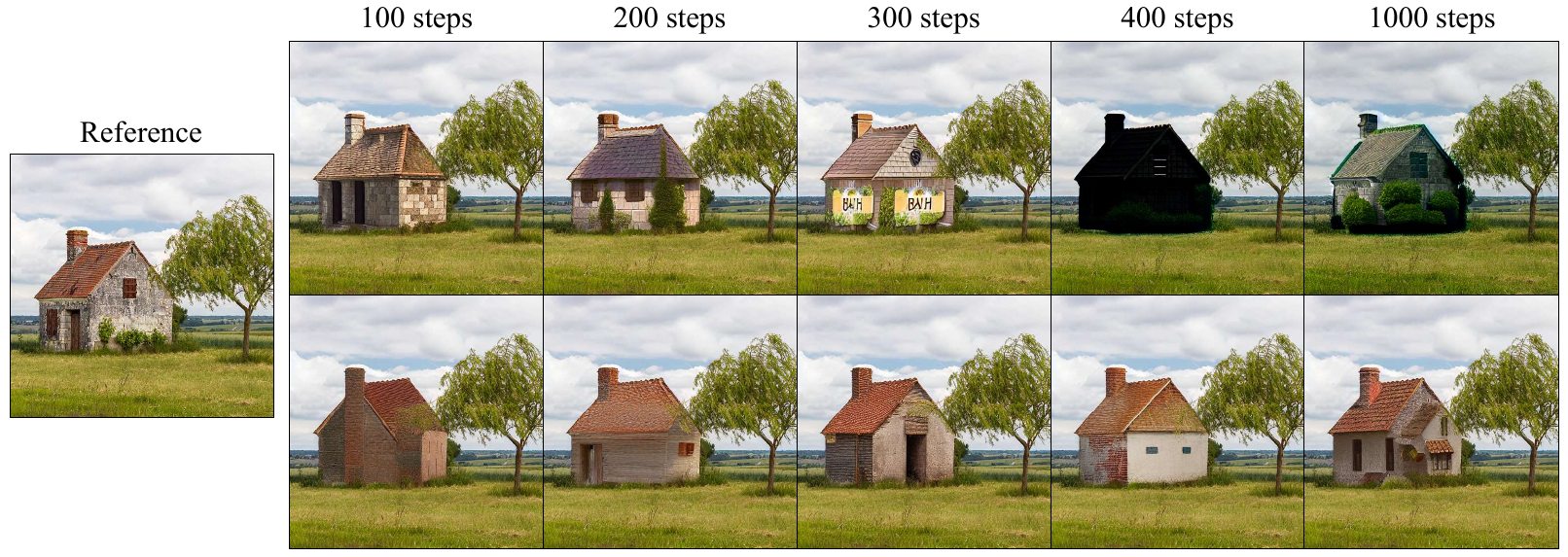}
\caption{The textual optimization can be executed both purely continuously or with discrete projection alternation. In the first row, we show the generation of images with the optimized text embedding at respective timesteps utilizing the diffusion's inpainting objective function. The second row displays generation with discretely projected optimization.}
\label{fig:discrete_ablation}
\vspace{-8pt}
\end{figure*}

To cover a wide range of inputs for image immunization, optimization stability is fundamental. One contributing factor is the text embedding optimization stage, where yielding a high-fidelity centroid results in effective optimization onward. As presented in our work, one can choose from optimizing fully continuously or in a semi-continuous manner with alternated token discretization. As originally motivated, we treat the projection to the token manifold as a regularization stage. Because the token embedding matrix represents a bag of primal semantic values (prior to the text encoder), the projection intercepts the adversarial compounding and continuously defaults any spurious features learned throughout the optimization. In Fig.~\ref{fig:discrete_ablation}, it is seen that while the continuous optimization approach (first row) quickly leads to overfitting and finds adversarial solutions, the projected method displays stable feature convergence. While the token projection carries its inherent instabilities due to its discrete update, the diffusion-based inpainting objective is naturally easier than the pure diffusion objective due to the provided hint via the context image, thus limiting the solution's search space.     

\newcolumntype{g}{>{\columncolor{Gray}}c}
\begin{table*}[h]
    \centering
    \caption{Following the same settings as Table 1 in the main paper, we compare the quantitative metrics between DDD and its ablated version DDD-A. Namely, DDD-A utilizes $\phi(\o)$ for the semantic digression. 
    }
  \label{tab:metrics_ablation}
    \begin{tabular}{lccccccll}
    \toprule
     &CLIP Score $\downarrow$  &Aesthetic $\downarrow$ & PickScore$\downarrow$& FID $\uparrow$ &  KID $\uparrow$&LPIPS $\uparrow$&SSIM $\downarrow$&PSNR$\downarrow$\\
    \hline
    (R $\rightarrow$ R) DDD-A& 0.2458&5.3946& 0.51& 111.43& 0.016434&0.4892& 0.5243&29.5674\\
    \rowcolor[gray]{0.9}
    (R $\rightarrow$ R) DDD& \B0.2443&\B5.2246&\B0.49&\B118.97&\B0.020934&\B0.4993&\B 0.5153&\B29.3763\\
    
    \hline
    (R $\rightarrow$ S) DDD-A& 0.2746& 5.5356& 0.52& 74.62& 0.002871& 0.3741& 0.6401& 29.9168\\
    \rowcolor[gray]{0.9}
    (R $\rightarrow$ S) DDD 
    & \B0.2739& \B5.4194&\B 0.48&\B 80.538&\B 0.004812&\B0.4032&\B0.6129&\B29.6729\\
    
    \hline
    (S $\rightarrow$ S) DDD-A& 0.2516&5.3804&\B 0.47& \B112.11& \B0.014769& 0.4342& 0.5842& 29.6690\\
    \rowcolor[gray]{0.9}
    (S $\rightarrow$ S) DDD &\B 0.2507& \B5.2805&0.53& 106.126& 0.013449&\B0.4423&\B 0.5751&\B29.5184\\
    
    \hline
    (S $\rightarrow$ R) DDD-A& 0.2787&5.5598& 0.52& 66.64& 0.001771&0.3915& 0.6043&29.8960\\
    \rowcolor[gray]{0.9}
    (S $\rightarrow$ R) DDD &\B0.2741&\B5.5284&\B 0.48& \B72.62&\B 0.003124& \B0.4084&\B 0.5859& \B29.6683\\
    
    \hline
    Oracle (R)&0.2582  &5.8147&-&-&-&-& -&-\\
    Oracle (S)&0.2597  &5.8168&-&-&-&-& -&-\\
   \bottomrule
   \end{tabular}
\end{table*}

\section{Metrics overview}\label{appendix:Metrics}
CLIP score \cite{hessel2021clipscore} measures the cosine alignment between the CLIP's image and text embedding. With the disrupted image and its prompted text, we calculate the mean CLIP score, where a lower score means misalignment between the two modalities and, thus, disruption. While the CLIP score measures the presence of features in the image, it may not consider how visually coherent they are synthesized across the image. The CLIP aesthetic predictor \cite{schuhmann2022laion} is a classifier trained on human evaluation to determine the aesthetics score of an image. These predictors are used to finetune generative models to suppress the generation of incoherent images. Recently, PickScore ~\cite{kirstain2023pick}, a human preference scoring network, has been developed for evaluating images and propagating network feedback. 
We use this score as a proxy of image disruption. 

Also, Structural Similarity Index Metric (SSIM)~\cite{1284395} quantifies the structural element in the image. We calculate between the natural and the disrupted inpainting, where lower scores mean greater disparity and disruption. PSNR measures the peak signal-to-noise ratio, in decibels, between two images. FID \cite{lucic2018gans} and KID \cite{binkowski2018demystifying} scores summarize the distance and the Maximum Mean Discrepancy, respectively, between the Inception \cite{szegedy2016rethinking} feature vectors of two datasets. Here, higher scores indicate dissimilarity between the natural and disrupted inpainting. Lastly, LPIPS score \cite{8578166} computes the distance between patch-wise activations of some pre-defined network. Here, we again compare between the natural and disrupted inpainting datasets.

\section{Human Evaluation}\label{appendix:humaneval}

\begin{figure*}
    \begin{center}
    \includegraphics[width=0.75\linewidth]{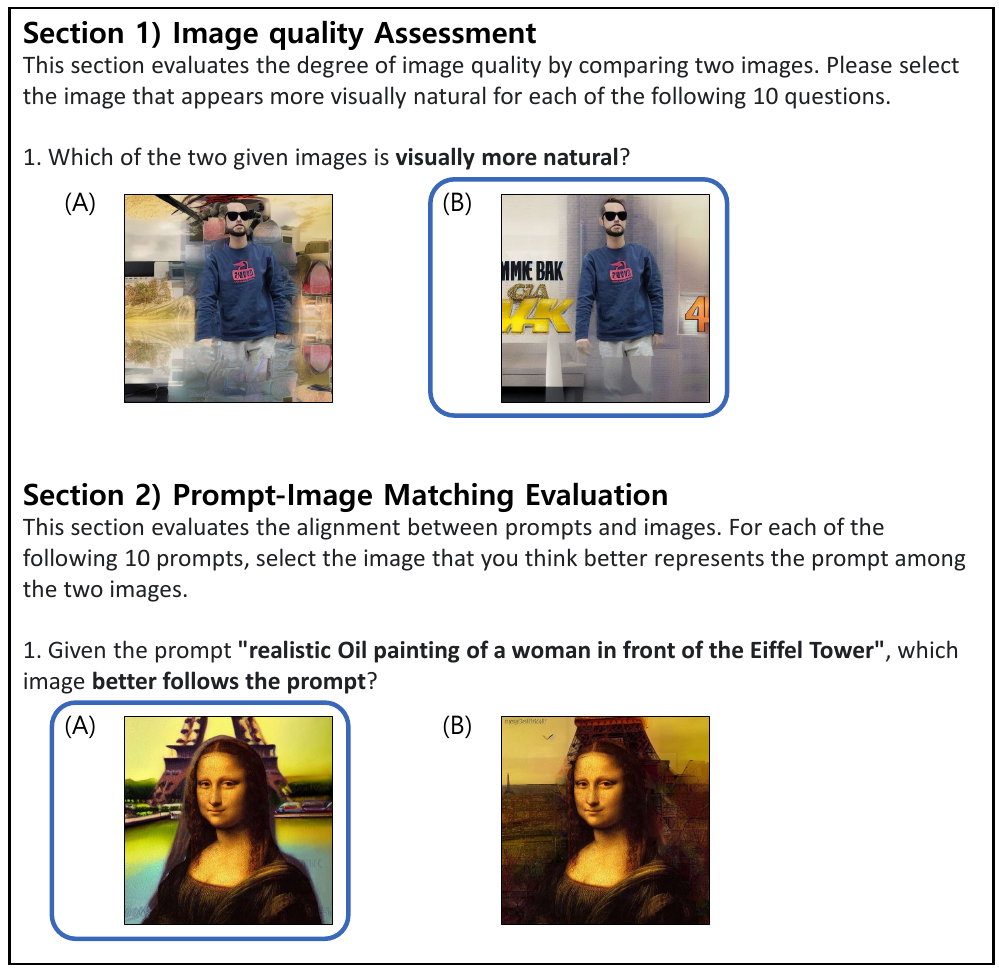}
    \end{center}
    \caption{Example of human evaluation for deepfake disruption.}
\label{fig:detail_survey}
\end{figure*}

For the collection on human preference over which disruption is more successful, we have surveyed 31 individuals via Google Forms. While our primary interest is gathering preferences on the disruption, choosing the "best disruption" is rather unintuitive and unnatural to most people. To this end, we formulate the questionnaire strategically to minimize the error from misunderstanding. Also, to both avoid individuals' entangled evaluation criteria and maximize the information, we decompose the disruption evaluation into the visual and semantic aspects. Fig. ~\ref{fig:detail_survey} shows key components of our survey.

\begin{figure*}
    \begin{center}
    \includegraphics[width=1.0\linewidth]{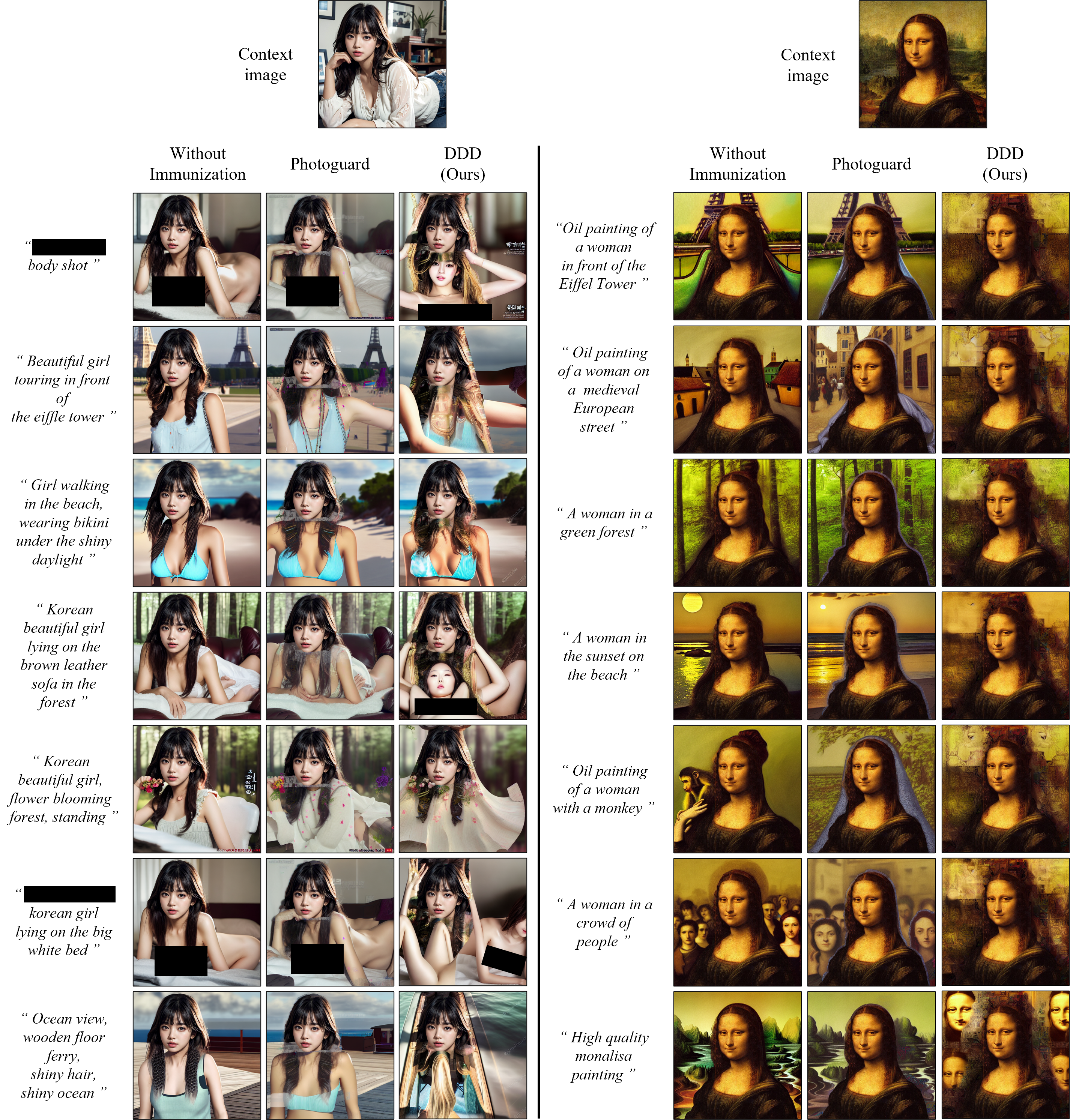}
    \end{center}
    \caption{Images of ``girl'' and ``monalisa'' generated and disrupted by Photoguard and DDD. We have censored violent/inappropriate words.}
\label{fig:detail_girl}
\end{figure*}

\begin{figure*}
    \begin{center}
    \includegraphics[width=\linewidth]{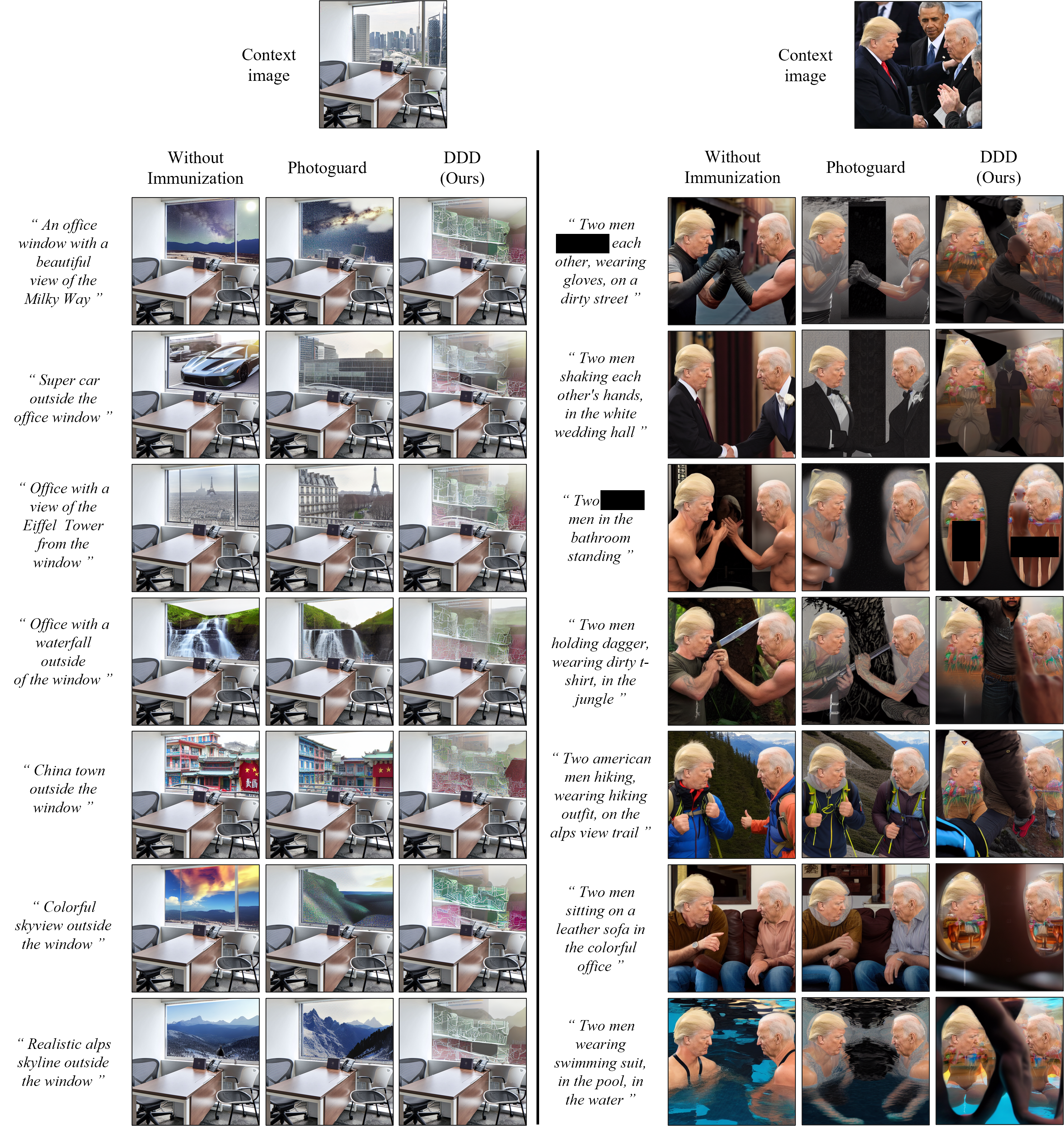}
    \end{center}
    \caption{Images of ``office'' and ``president'' generated and disrupted by Photoguard and DDD. We have censored violent/inappropriate words.}
\label{fig:detail_house}
\end{figure*}